\documentclass[letterpaper, 10 pt, conference]{ieeeconf}  

\IEEEoverridecommandlockouts                              

\usepackage{graphicx, cite, bm}
\usepackage{amsmath}
\usepackage{xcolor}
\usepackage{amsfonts}
\usepackage{multirow}
\usepackage{mathtools}
\usepackage[cjk]{kotex}
\usepackage{siunitx} 
\usepackage{tabularx}
\usepackage[caption=false,font=footnotesize]{subfig}

\usepackage{dblfloatfix}
\usepackage[hidelinks]{hyperref}
\usepackage{makecell}
\usepackage{siunitx}
\usepackage{threeparttable}
\usepackage[normalem]{ulem}
\usepackage{booktabs} 
\usepackage{bbm}      

\title{\LARGE \bf 
Design of a 3-DOF Hopping Robot with an Optimized Gearbox: \\ An Intermediate Platform Toward Bipedal Robots
}

\author{JongHun Choe$^{1}$, Gijeong Kim$^{1}$, Hajun Kim$^{1}$, Dongyun Kang$^{1}$, Min-Su Kim$^{1}$ and Hae-Won Park$^{1}$, \textit{Member, IEEE}
\thanks{
This work was supported by the Technology Innovation Program(or Industrial Strategic Technology Development Program-Robot Industry Technology Development)(00427719, Dexterous and Agile Humanoid Robots for Industrial Applications) funded By the Ministry of Trade Industry \& Energy(MOTIE, Korea).
}
\thanks{
$^{1}$Authors are with the Humanoid Robot Research Center, School of Mechanical, Aerospace \& Systems Engineering, Department of Mechanical Engineering, Korea Advanced Institute of Science and Technology (KAIST), Yuseong-gu, 34141 Daejeon, Republic of Korea. {\tt\small haewonpark@kaist.ac.kr}}
}

\begin{document}

\maketitle
\thispagestyle{empty}
\pagestyle{empty}

\begin{abstract}

This paper presents a 3-DOF hopping robot with a human-like lower-limb joint configuration and a flat foot, capable of performing dynamic and repetitive jumping motions.
To achieve both high torque output and a large hollow shaft diameter for efficient cable routing, a compact 3K compound planetary gearbox was designed using mixed-integer nonlinear programming for gear tooth optimization.
To meet performance requirements within the constrained joint geometry, all major components—including the actuator, motor driver, and communication interface—were custom-designed.
The robot weighs 12.45~\si{\kilogram}, including a dummy mass, and measures 840~\si{\milli\meter} in length when the knee joint is fully extended.
A reinforcement learning-based controller was employed, and the robot’s performance was validated through hardware experiments, demonstrating stable and repetitive hopping motions in response to user inputs.
These experimental results indicate that the platform serves as a solid foundation for future bipedal robot development.
A supplementary video is available at: \url{https://youtu.be/BZ2H0dQBcXc}
\end{abstract}

\section{Introduction} \label{Sec:Introduction}
Dynamic legged locomotion remains a core challenge in humanoid robotics, requiring actuators to provide both high torque for impulsive ground interaction and low inertia for agile movement.
Among various locomotion strategies, hopping is attractive due to its ability to achieve dynamic and agile motion with a simple mechanical configuration and a small number of actuators, while also presenting complex control challenges such as managing intermittent contact and flight phase~\cite{sayyad2007single}.
Due to these difficulties, hopping robots serve as a valid approach for studying agile legged locomotion and act as a practical intermediate prototype for validating actuators and control strategies.

Many existing hopping robots are based on point-foot contact models~\cite{raibert1986legged,franccois1998new,cherouvim2005single,ahmadi2006controlled,hurst2010actuator,an2022design}, which simplify foot-ground interaction and reduce control complexity.
While effective for demonstrating basic hopping motions, these models lack the spatial constraints and contact dynamics associated with flat-footed humanoid legs. 
Some systems introduce compliance or articulated toes~\cite{hyon2002development,mombaur2002stable}, aiming to improve walking stability, posture control, and dynamic performance.
However, to the best of the authors' knowledge, repeated flat-foot hopping with a human-like joint layout remains largely unexplored in both the literature and hardware demonstrations.

\begin{figure}[!t]
  \centering
  \includegraphics[width=\columnwidth]{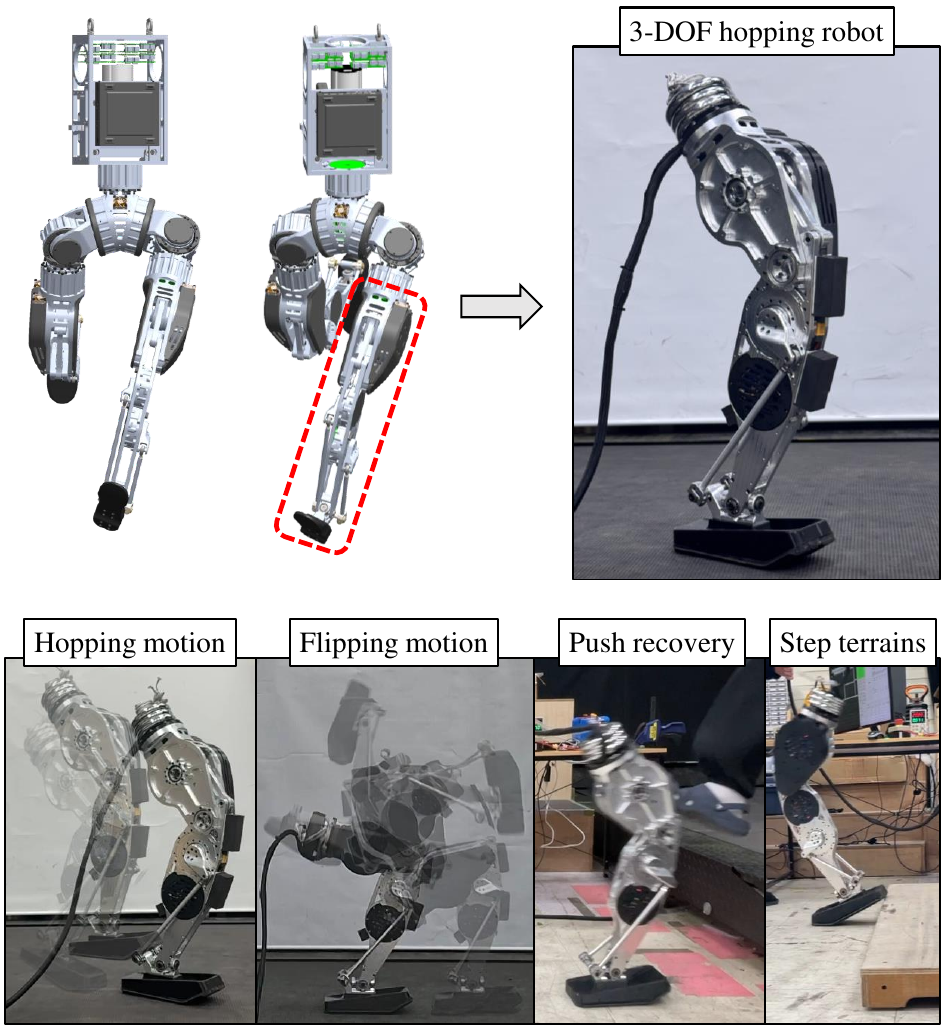}
  \caption{Conceptual design of the bipedal robot currently under development and the proposed 3-DOF hopping robot as an intermediate prototype. The lower images show the proposed robot performing hopping and flipping motions, and responding to external push disturbances and step terrains.}
  \label{fig:figure_1}
\end{figure}

In this work, we present a 3-DOF hopping robot with a flat foot and human-like joint configurations, consisting of one knee DOF and two ankle DOFs. Designed as an intermediate prototype for a full bipedal robot platform with 13 DOF (6 DOF per leg and 1 DOF at the waist for yaw), as illustrated in Figure~\ref{fig:figure_1}, the robot demonstrates dynamically stable hopping, front flips, and push recovery, exhibiting robust performance against terrain variations such as step changes.

To enable this dynamic behavior, an actuator that can deliver both high torque and high velocity is required.  
To achieve these performance requirements, we selected the quasi-direct drive (QDD) concept among various actuation strategies, which combines high torque-density motors with low gear ratios.  
QDD has been widely used in dynamic legged robots due to its high torque transparency and backdrivability~\cite{seok2014design,bledt2018cheetah,park2017high}.  
However, to meet the high torque demands at the humanoid scale within the limited joint space, a higher gear ratio is required, and a hollow shaft structure is also necessary to ensure reliable internal cable routing.  
To address these challenges, we adopted a 3K compound planetary gearbox as the reducer.  
From a design optimization perspective, mixed-integer nonlinear programming (MINLP) techniques have shown promise in synthesizing gear trains under tight physical constraints~\cite{shin2022design}.  
Accordingly, the gear tooth configuration was optimized using an MINLP method tailored to the motor’s torque-speed envelope and the geometric constraints of the joint assembly.

The hardware system was developed with custom-designed major components, including the actuator, motor driver, and communication interface. A learning-based controller was employed to enable dynamic behaviors such as repetitive hopping and push recovery. While the prototype presented in this paper constitutes only a partial subsystem of a planned full bipedal robot, the experimental results demonstrate actuator performance and structural robustness. These results suggest that the platform provides a suitable basis for future bipedal development.

The remainder of this paper is organized as follows: Section~\ref{Sec:Design} presents the hardware design of the proposed hopping robot.
Section~\ref{Sec:Control} describes the reinforcement learning-based control policy.
Section~\ref{Sec:Experiments} validates the performance of the proposed hopping robot through hardware experiments.
Finally, Section~\ref{Sec:Conclusion} concludes the paper.
\section{Hardware Design} \label{Sec:Design}

\begin{figure}[!t]
  \centering
  \includegraphics[width=\columnwidth]{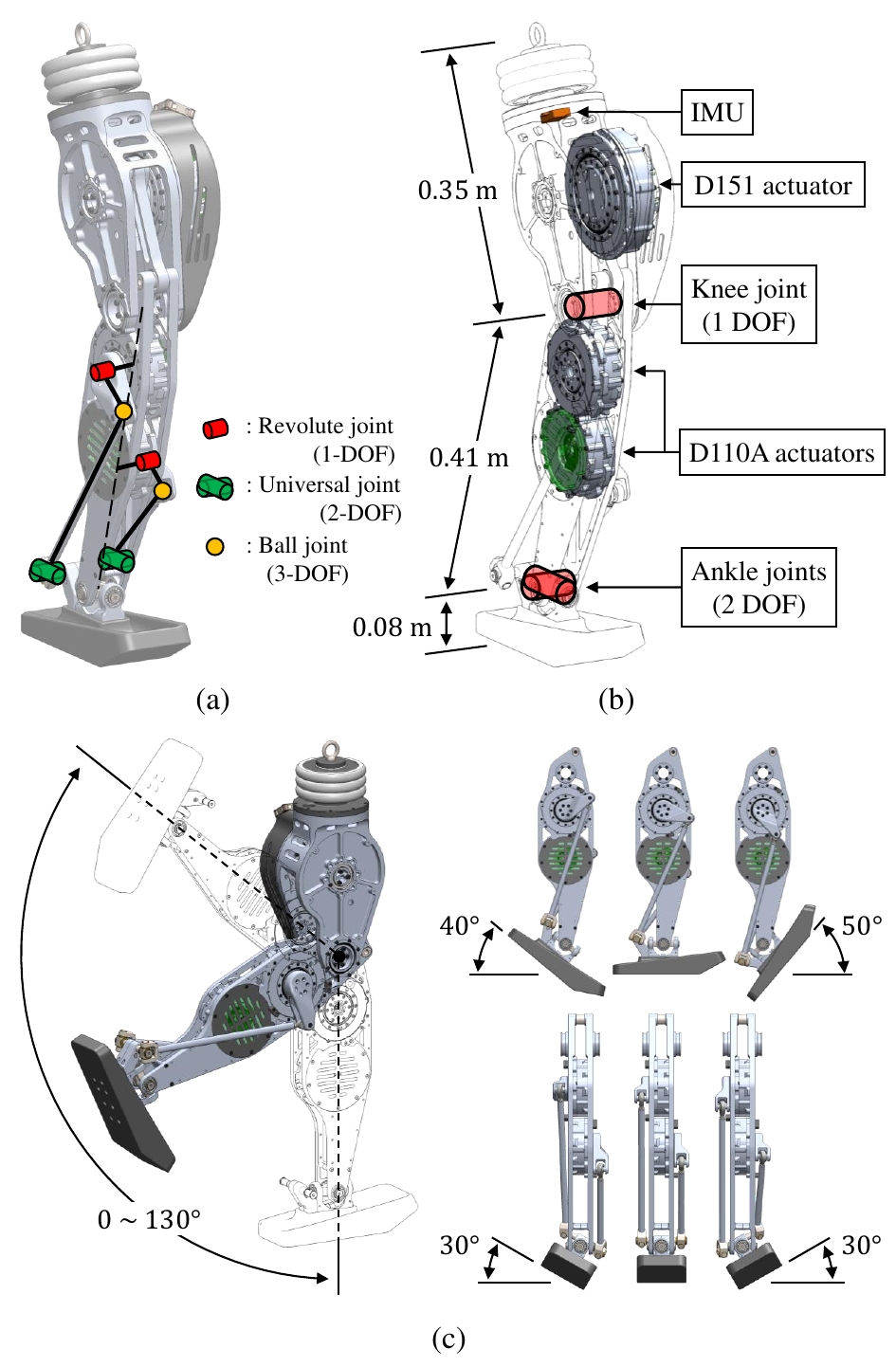}
  \caption{Overview of the proposed hopping robot: (a) design and schematic of the closed-loop parallel mechanism in the ankle joint, (b) key dimensions, major components, and joint configurations, and (c) workspace of the robot.}
  \label{fig:robot_overview}
\end{figure}

\subsection{Mechanical Structure}
The three-dimensional design, key dimensions, major components, and joint configuration of the hopping robot proposed in this paper are shown in Fig.~\ref{fig:robot_overview}~(a) and (b).
The robot's joint configuration is inspired by the lower limb of the human leg, and it consists of a total of 3-DOF, including a 1-DOF knee joint and 2-DOF ankle joints.
Two types of custom-designed actuators are implemented, where the D151 actuator is assigned to the knee joint and the D110A actuator is assigned to the ankle joints.
A more detailed description of these two actuators is provided in a later section.
All actuators are positioned close to the upper link of their corresponding joint in order to reduce the rotational inertia.

The power transmission mechanism is designed differently for each joint.
For the knee joint, torque is transmitted through a four-bar linkage with a 1:1 transmission ratio.
For the ankle pitch and roll joints, a closed-loop parallel mechanism, as shown in Fig.~\ref{fig:robot_overview}~(a), is employed~\cite{jeong2020design}.
A universal joint is implemented at the foot link, and a ball joint is incorporated at the output link of the D110A actuators.
This design allows the ankle to achieve two degrees of freedom (pitch and roll) motion, even though the output shafts of the two D110A actuators are placed in parallel.

The proposed hopping robot has a total mass of 12.45~\si{\kilo\gram} and a full extension length of 890~\si{\milli\meter} when the knee joint is fully extended.
Table~\ref{tab:segment-comparison} compares the length and mass of each segment of the robot with human data, based on an adult male with a height of 165~\si{\centi\meter} and a weight of 75~\si{\kilo\gram}.
Overall, most of the segment dimensions are similar, although the thigh segment shows a small difference because the hip structure has not been fully implemented, and a dummy mass was used instead.

The range of motion for each joint of the proposed robot is shown in Fig.~\ref{fig:robot_overview}~(c).
The ankle pitch joint moves from -50~\si{\degree} to 40~\si{\degree}, the ankle roll joint can be actuated to move from -30~\si{\degree} to 30~\si{\degree}, and the knee joint has a range of motion from 0~\si{\degree} to 130~\si{\degree}.
The knee joint is designed to allow full extension, enabling the implementation of human-like gait motions such as straight-knee walking when the robot is expanded into a bipedal walking robot.

\begin{table}[!t]
\renewcommand*{\arraystretch}{1.3}
\centering
\caption{Robot and Human Segment Data Comparison}
\label{tab:segment-comparison}
\begin{tabular}{lcc|cc}
\hline
\textbf{} & \multicolumn{2}{c|}{Mass [kg]} & \multicolumn{2}{c}{Length [m]} \\
\hline
\textbf{} 
& Robot 
& Human (\%) 
& Robot 
& Human (\%) \\
\hline
Thigh & 7.60 & 7.86 (10.5) & 0.35 & 0.383 (23.2) \\
Calf  & 3.89 & 3.56 (4.75) & 0.41 & 0.408 (24.7) \\
Foot  & 0.96 & 1.07 (1.43) & 0.08 & 0.070 (4.25) \\
\hline
\end{tabular}
\end{table}

\subsection{Electrical Architecture} \label{sec:electrical_architecture}

\begin{figure}[!t]
  \centering
  \includegraphics[width=\columnwidth]{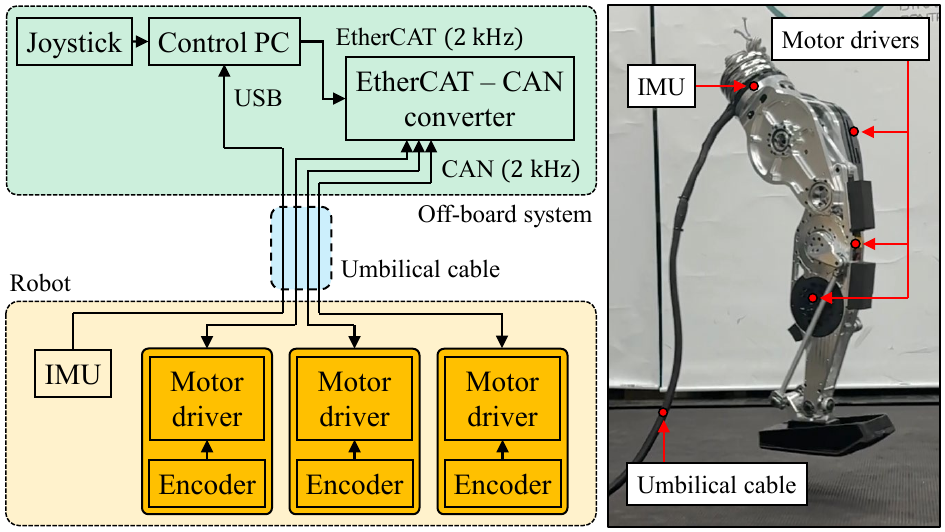}
  \caption{Electrical architecture of the proposed hopping robot}
  \label{fig:electrical_architecture}
\end{figure}

The electrical architecture of the proposed hopping robot is illustrated in Fig.~\ref{fig:electrical_architecture}.
The overall system consists of a robot system, an off-board system, and an umbilical cable connecting the two.
A tethered system configuration was adopted because there was insufficient space in the robot’s hip section to mount all the electronic components and battery pack.

Although the power lines are omitted from the diagram, the off-board system uses external power.
The robot system is powered by a battery pack consisting of 18650 30T battery cells arranged in a 16S4P configuration, and this battery pack supplies power to each actuator through the umbilical cable.

User commands are sent to the control PC via Bluetooth using a Logitech F10 joystick.
Control computations are performed on a ThinkStation P360 Tiny equipped with an Intel i7-11700T processor.
The computed control inputs are transmitted over EtherCAT protocol to a custom-designed EtherCAT–CAN converter, which converts the signals to CAN protocol and transmits them to the robot system through the umbilical cable.
Both the EtherCAT and CAN operate at a frequency of 2~\si{\kilo\hertz}.
Since the robot performs highly dynamic motions such as hopping, CAN protocol was selected to ensure reliable transmission of control signals to the actuators.

An RLS MB039 encoder and a MicroStrain 3DM-GV7-AR IMU were employed to measure the joint states and estimate the base orientation, respectively.

\begin{table}[!t]
\renewcommand*{\arraystretch}{1.3}
\centering
\begin{threeparttable}
\caption{Actuator Specifications}
\label{tab:d151-d110a}
\begin{tabular}{>{\raggedleft\arraybackslash}m{3.0cm}|>{\centering\arraybackslash}m{2.1cm}>{\centering\arraybackslash}m{2.1cm}}
\hline
 & D151 & D110A \\
\hline
Usage & Knee & Ankle \\
Gearbox type & 3K compound planetary gearbox & Single stage planetary gearbox \\
Gear ratio & $20:1$ & $8:1$ \\
\makecell[r]{Peak torque \\ @ 50~\si{\ampere}~(\si{\newton\meter})} & $320$ & $176$ \\
\makecell[r]{Peak angular velocity \\ @ 67.2~\si{\volt}~(\si{\radian/\second})} & $10$ & $20$ \\
Torque constant $(\mathrm{Nm/A})$ & $0.32$ & $0.44$ \\
Dimension $(\mathrm{mm})$ & $\mathrm{\Phi}\, 170 \times 47.2$\tnote{*} & $\mathrm{\Phi}\, 131 \times 40.5$\tnote{*} \\
Mass $(\mathrm{kg})$ & $2.27$ & $1.1$ \\
Rotor inertia $(\mathrm{kg\cdot m^2})$ & $0.000922$\tnote{*} & $0.0002$\tnote{*} \\
\hline
\end{tabular}
\begin{tablenotes}
\item[*] Measured in the three-dimensional design software.
\end{tablenotes}
\end{threeparttable}
\end{table}

\subsection{Custom-Designed Components}

The hopping robot proposed in this paper features custom-designed major hardware components including the actuators (motor and gearbox), motor drivers, and the EtherCAT–CAN converter, to achieve the required performance while meeting the structural constraints.

\begin{figure}[!t]
  \centering
  \includegraphics[width=\columnwidth]{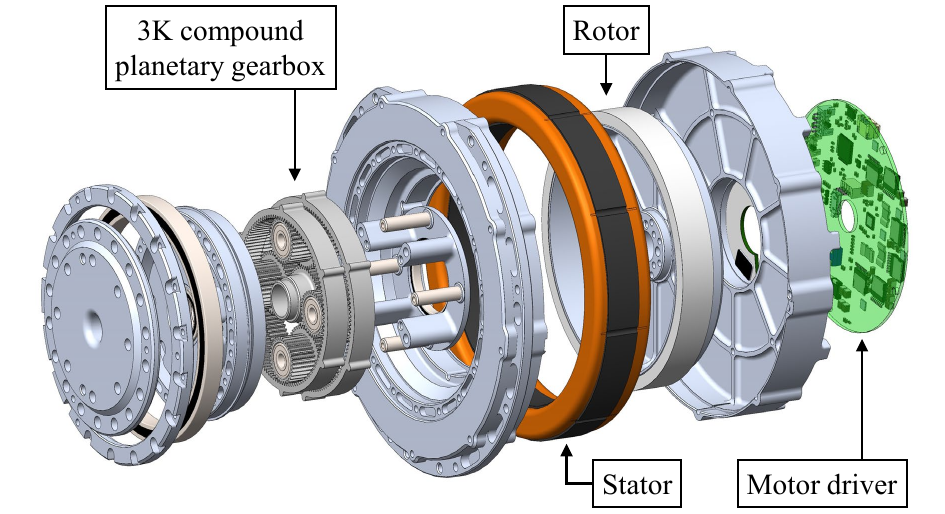}
  \caption{Exploded view of the D151 actuator for the knee joint.}
  \label{fig:exploded_view}
\end{figure}

Fig.~\ref{fig:exploded_view} presents the exploded view of the D151 actuator and shows its major components, including the motor (stator and rotor), a gearbox, and a motor driver.
The specifications of the D151 and D110A actuators are summarized in Table~\ref{tab:d151-d110a}.
To meet the torque and angular velocity requirements while accommodating the limited available space at each joint, custom-designed motors, gearboxes, and motor drivers were developed.

\begin{figure}[!t]
  \centering
  \includegraphics[width=\columnwidth]{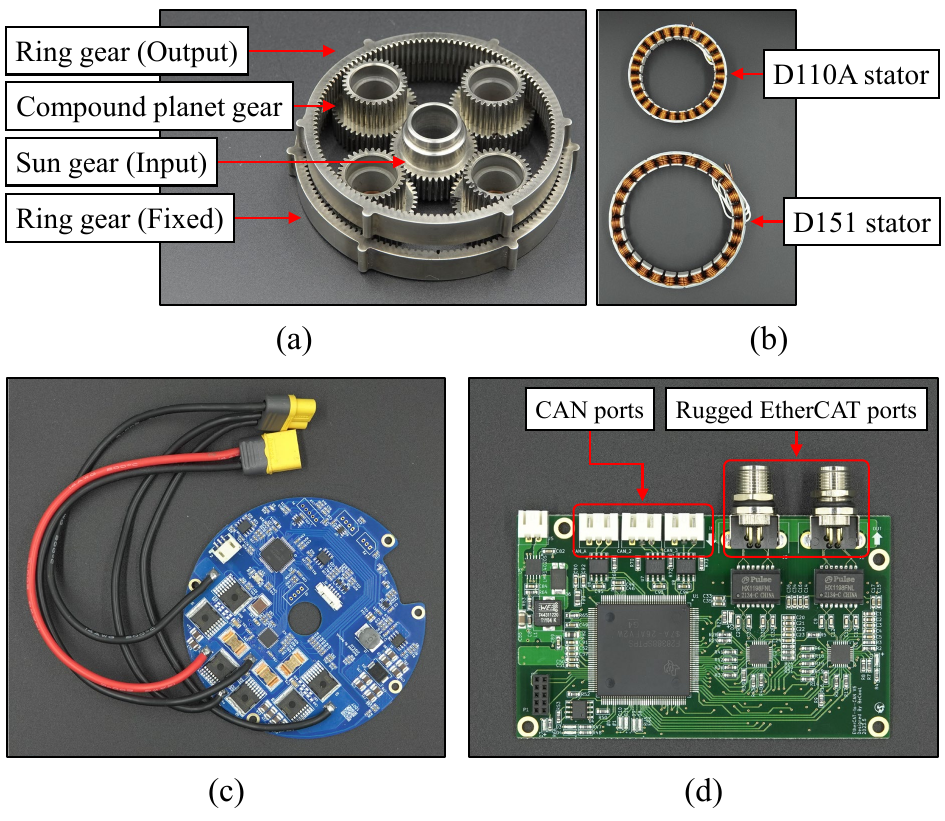}
  \caption{Photographs of the custom-designed components of the proposed hopping robot: (a) 3K compound planetary gearbox used in the D151 actuator, (b) D110A and D151 motor stators, (c) motor driver, and (d) EtherCAT-CAN converter.}
  \label{fig:custom_designed_components}
\end{figure}

The custom-designed D151 and D110A stators are shown in Fig.~\ref{fig:custom_designed_components}~(b).
The stators were fabricated using VACOFLUX48 cobalt-based steel sheets to achieve high torque density, and the rotors were equipped with neodymium magnets of 52SH grade.
The gearboxes were designed through a gear teeth optimization process based on MINLP, which is described in the following section.
The number of teeth for each gear was selected to maximize the inner diameter of the hollow shaft to facilitate wiring, while avoiding interference with the frame structure and reducing the overall gearbox mass.

The motor drivers of the proposed robot are mounted directly behind each actuator to enable wiring, thus affecting the overall actuator size.
Custom-designed motor drivers were developed to fit within the compact space while achieving the required operating voltage and current control capability.
As shown in Fig.~\ref{fig:custom_designed_components}~(c), the motor drivers support an input voltage of up to 100~\si{\volt} and a current control frequency of 20~\si{\kilo\hertz}.

The EtherCAT–CAN converter, which was custom-designed as described above, is illustrated in Fig.~\ref{fig:custom_designed_components}~(d).
It features rugged EtherCAT input and output ports to ensure robust communication and employs a TMS320F28388 microcontroller for control signal conversion.
The converted CAN signals are then transmitted via three CAN ports to the motor drivers responsible for each actuator in the robot system.

\begin{figure}[!t]
  \centering
  \includegraphics[width=\columnwidth]{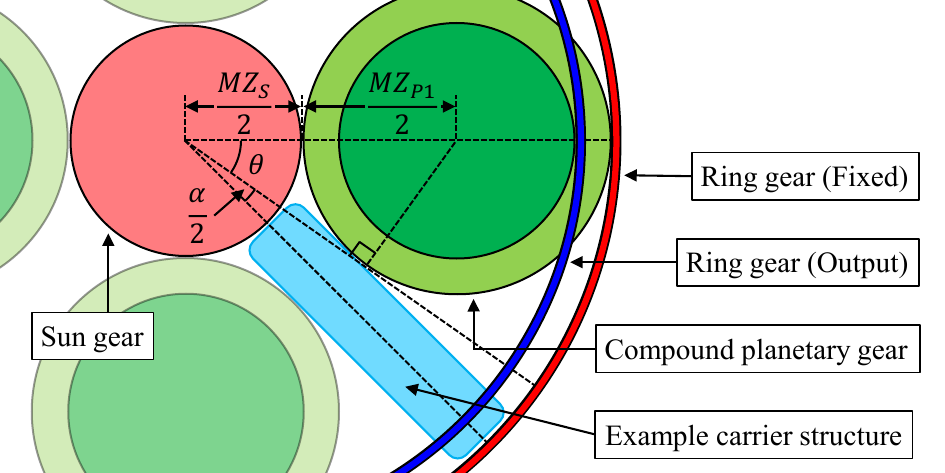}
  \caption{Geometric relationship between the compound planet gears and the carrier structure in the 3K compound planetary gearbox.}
  \label{fig:carrier_structure}
\end{figure}

\subsection{Gear Teeth Optimization of 3K Compound Planetary Gearbox}

The knee joint of the proposed robot employs a 3K compound planetary gearbox rather than a conventional single-stage planetary gearbox.
Since the knee joint requires high torque performance, it demands a high gear ratio while simultaneously realizing a compact structure within the limited joint space and incorporating a hollow shaft that facilitates wiring.

The 3K compound planetary gearbox is composed of five distinct types of gears, where the number of teeth for each gear must satisfy both the target gear ratio and physical interference constraints.
Manual selection of the gear teeth numbers is challenging and time-consuming, particularly due to the nonlinear dependence of the gear ratio on the gear combination, which complicates the systematic exploration of the design space.

Therefore, in this work, MINLP was employed to select the gear teeth numbers, ensuring that the gear geometry and physical constraints are satisfied while achieving the target gear ratio.

\subsubsection{Optimization Variables}
The variables used in this optimization problem are the number of teeth of each gear constituting the gearbox, all of which are defined as positive integers:
\begin{itemize}
    \item \( Z_{S} \): Number of teeth of the sun gear
    \item \( Z_{P1} \): Number of teeth of the planet gear (Input)
    \item \( Z_{P2} \): Number of teeth of the planet gear (Output)
    \item \( Z_{F} \): Number of teeth of the fixed ring gear
    \item \( Z_{O} \): Number of teeth of the output ring gear
\end{itemize}
Since a compound planet gear is used, there exist two types of planet gears with $Z_{P1}$ and $Z_{P2}$ teeth, respectively.
The input planet gear meshes with both the sun gear and the fixed ring gear, while the output planet gear engages with the output ring gear.

\subsubsection{Cost Function}
The objective of this optimization problem is to maximize the diameter of the hollow shaft by increasing the diameters of the sun gear and the fixed ring gear, while minimizing the overall mass of the gearbox by reducing the diameters of the other gears.
Accordingly, the cost $J$ is defined as follows:
\begin{equation} \label{eqn:cost_function}
    J = \left( \frac{1}{Z_{S}} \right)^2 + \left(Z_{F} - \frac{D_{}}{M} \right)^2 + Z_{P1}^2 + Z_{P3}^2 + Z_{O}^2
\end{equation}
where $D$ is the inner diameter of the rotor and $M$ is the module of the gears.
This approach enables the gearbox to have an outer diameter smaller than the rotor’s inner diameter, allowing it to be housed inside the rotor and thereby reducing the overall height of the actuator.

\subsubsection{Equality Constraints}
The equality constraints used in this optimization problem can be classified into structural feasibility conditions, assembly feasibility conditions, and conditions related to the gear ratio.
First, the structural feasibility conditions of the gearbox are defined as follows:
\begin{align}
    Z_{F} &= Z_{S} + 2 Z_{P1} \label{eqn:equality_constraint_1} \\
    Z_{O} &= Z_{S} + Z_{P1} + Z_{P2} \label{eqn:equality_constraint_2}
\end{align}

Next, the conditions for the assembly are formulated as follows~\cite{Zou2015Single}:
\begin{equation}
    \frac{Z_{F}+Z_{S}}{n_{P}} \in \mathbb{Z}, \quad \frac{2Z_{O}-2Z_{P2}}{n_{P}} \in \mathbb{Z} \label{eqn:equality_constraint_3}
\end{equation}
where $n_{P}$ denotes the number of compound planet gears, and $\mathbb{Z}$ represents the set of integers.

Lastly, the condition related to the gear ratio is given as follows:
\begin{equation}
    G_{target} = \frac{2Z_{P1}(Z_{F}-Z_{P1}+Z_{P2})}{(Z_{F}-2Z_{P1})(Z_{P1}-Z_{P2})} \label{eqn:equality_constraint_4}
\end{equation}
where $G_{\text{target}}$ denotes the target gear ratio, which is set to 20 for the D151 actuator used in this study.

\subsubsection{Inequality Constraints}
The inequality constraints of the proposed optimization problem are formulated to satisfy mechanical and physical design requirements.

The first constraint addresses the relationship between the number of compound planet gears and the carrier structure, considering potential interference between them.
As illustrated in Fig.~\ref{fig:carrier_structure}, this relationship is expressed as follows:
\begin{equation}
    2 n_{P} \left( \theta + \frac{\alpha}{2} \right) = 2\pi, \quad \theta=\arcsin \left( \frac{Z_{P1}}{Z_{S}+Z_{P1}} \right) \nonumber
\end{equation}
where $\alpha$ represents the angular clearance available for the carrier structure.
By setting the minimum allowable clearance for the carrier structure, $\alpha_{\min}$,
as a hyperparameter, the following constraint is derived:
\begin{equation}
    \alpha_{\min} \leq \frac{\pi}{n_{P}} - \arcsin \left( \frac{Z_{P1}}{Z_{S}+Z_{P1}} \right) \label{eqn:inequality_constraint_1}
\end{equation}

\begin{table}[!t]
\centering
\caption{Barrier Reward Terms}
\label{tab:reward_barrier}
\scriptsize
\begin{threeparttable}
\begin{tabular}{@{}lllll@{}}
\toprule
\textbf{Term} & \textbf{Constraint Variable} & $d^{\text{lower}}$$^{\ast}$ & $d^{\text{upper}}$$^{\ast}$ & $\delta$$^{\ast}$ 
 \\ \midrule
$b_{\text{gait}}$ & Gait, $f_i$ & $-0.3$ & $2.0^{\dagger}$ & $0.08$ \\
$b_{\text{foot}}$ & Foot clearance, $l_i$ [m] & $-0.08$ & $1.0^{\dagger}$ & $0.02$ \\
$b_{\text{body}}$ & Body height, $h$ [m] & $0.30$ & $1.10$ & $0.04$ \\
$b_{\text{cmd}}$ & 
\begin{tabular}[c]{@{}l@{}}Command velocity tracking, \\ 
$v_x - v_x^{\text{cmd}}$, $v_y - v_y^{\text{cmd}}$ [m/s], \\ $\omega_z - \omega_z^{\text{cmd}}$ [rad/s]
\end{tabular} &
$\begin{bmatrix}-0.6 \\ -0.6 \\ -0.6\end{bmatrix}$ & 
$\begin{bmatrix}0.6 \\ 0.6 \\ 0.6\end{bmatrix}$ & 
$\begin{bmatrix}0.6 \\ 0.6 \\ 0.6\end{bmatrix}$ \\
$b_{\text{pos}}$ & \begin{tabular}[c]{@{}l@{}}Joint position, $q$ [rad] \\ (knee, ankle pitch, ankle roll)\end{tabular} &
$\begin{bmatrix}0.11 \\ -0.74 \\ -0.38\end{bmatrix}$ &
$\begin{bmatrix}2.09 \\ 0.62 \\ 0.38\end{bmatrix}$ &
$\begin{bmatrix}0.08 \\ 0.08 \\ 0.08\end{bmatrix}$ \\
$b_{\text{vel}}$ & Joint velocity, $\dot{q}$ [rad/s] & $-8$ & $8$ & $2.0$ \\
$b_{\text{base}}$ & 
\begin{tabular}[c]{@{}l@{}}Base motion, $v_z$ [m/s], \\ $\omega_x$, $\omega_y$ [rad/s] \end{tabular} & 
$\begin{bmatrix}-1.2 \\ -0.8 \\ -1.2\end{bmatrix}$ & 
$\begin{bmatrix}1.2 \\ 0.8 \\ 1.2\end{bmatrix}$ & 
$\begin{bmatrix}1.0 \\ 0.6 \\ 1.0\end{bmatrix}$ \\
$b_{\text{contact}}$ & Body contact, $c_{\text{body}}$ & $-1.0$ & $1.0$ & $0.5$ \\
$b_{\text{com}}$ & \begin{tabular}[c]{@{}l@{}}COM XY offset, $\mathbf{x}_{\text{com}}^{xy}$ [m] \\ \text{(standing mode only)}\end{tabular} & 
$\begin{bmatrix}-0.04 \\ -0.02\end{bmatrix}$ & 
$\begin{bmatrix}0.04 \\ 0.02\end{bmatrix}$ & 
$\begin{bmatrix}0.02 \\ 0.01\end{bmatrix}$ \\
\bottomrule
\end{tabular}

\vspace{1ex}
\begin{minipage}{\linewidth}
\scriptsize
\textit{Notation:} 
$^{\ast}$~$d^{\text{lower}}$ and $d^{\text{upper}}$ denote the lower and upper bounds of each constraint variable, respectively, and $\delta$ controls the curvature steepness in the relaxed log-barrier function. $f_i$ is a sinusoidal gait-phase signal signed by foot contact. A lower bound on $f_i$ regulates stance and swing durations: values near $-1$ allow flexible timing, while those near $0$ enforce symmetry around half the gait period~\cite{kim2024learning}. $l_i$ denotes the difference between the desired and actual foot height during swing. 
$c_{\text{body}}$ counts the number of body contact points with the ground. 
$\mathbf{x}_{\text{com}}^{xy}$ is the lateral offset of the center of mass relative to the stance foot in the body frame. 
$(\cdot)^{\text{cmd}}$ denotes commanded quantities. 
{$^\dagger$}For $f_i$ and $l_i$, the upper bounds are unnecessary and set to non-reachable values (2.0 and 1.0, respectively).
\end{minipage}

\vspace{1ex}
\begin{minipage}{\linewidth}
\scriptsize
\textit{Barrier Reward Description:} Each constraint variable $z_i$ is shaped via a relaxed log-barrier function $\phi(z_i;d^{\text{lower}}_i, d^{\text{upper}}_i, \delta_i)$, which guides $z_i$ into the constraint range $[d^{\text{lower}}_i, d^{\text{upper}}_i]$. The penalty increases sharply outside the bounds, with the steepness controlled by $\delta_i$. The total barrier reward is computed as a weighted sum:
\begin{equation*}
r_{\text{barrier}} = \sum_i \gamma_i \cdot \phi(z_i;d^{\text{lower}}_i, d^{\text{upper}}_i, \delta_i),
\end{equation*}
where all weights $\gamma_i$ are assigned the same value. Please refer to~\cite{kim2024learning} for the functional form of the barrier function.
\end{minipage}
\end{threeparttable}
\end{table}

The next constraints are about the size of each gear.
Considering the diameter of the hollow shaft and the bearings used, the gear size constraints are defined as follows: 
\begin{equation}
    Z_{S} \geq Z_{S,\min}, \quad Z_{P1} \geq Z_{P1,\min}, \quad Z_{P2} \geq Z_{P2,\min} \label{eqn:inequality_constraint_2}
\end{equation}
\begin{equation}
    Z_{F} \leq \frac{D}{M} \label{eqn:inequality_constraint_3}
\end{equation}

Additionally, inequality constraints are formulated to define the size relationships between the two planet gears within the compound planet gear and between the fixed and output ring gears.
These constraints are given as follows:
\begin{equation}
    Z_{P1} - Z_{P2} \geq n_{P}, \quad Z_{F} - Z_{O} \geq n_{P} \label{eqn:inequality_constraint_4}
\end{equation}

Based on the optimization variables, cost function (\ref{eqn:cost_function}), equality constraints (\ref{eqn:equality_constraint_1}), (\ref{eqn:equality_constraint_2}), (\ref{eqn:equality_constraint_3}), (\ref{eqn:equality_constraint_4}), and inequality constraints (\ref{eqn:inequality_constraint_1}), (\ref{eqn:inequality_constraint_2}), (\ref{eqn:inequality_constraint_3}), (\ref{eqn:inequality_constraint_4}), the optimization problem proposed in this study can be formulated.
Since this problem involves integer optimization variables along with a nonlinear cost function and nonlinear constraints, it corresponds to a MINLP problem.
In this paper, BARON was employed as the solver to solve the optimization problem~\cite{Sahinidis1996BARON}.

As a result of the optimization, the selected number of gear teeth for the D151 actuator gearbox is as follows: $Z_{S}=44$, $Z_{P1}=44$, $Z_{P2}=32$, $Z_{F}=132$, and $Z_{O}=120$.
The actual 3K compound planetary gearbox fabricated based on these gear teeth selections is shown in Fig.~\ref{fig:custom_designed_components}~(a).
The inner diameter $D$ of the D151 rotor was set to 79.4~\si{\milli\meter}, and the module $M$ of the gears was set to 0.6, resulting in a hollow shaft diameter of 12~\si{\milli\meter}.

\begin{table}[!t]
\centering
\caption{Standard Reward Terms}
\label{tab:reward_standard}
\scriptsize
\begin{threeparttable}
\begin{tabular}{@{}lll@{}}
\toprule
\textbf{Term} & \textbf{Description} & \textbf{Expression} \\
\midrule
\multicolumn{3}{l}{\textit{Positive Rewards (to be maximized)}} \\
$r_{\text{lin}}$ & Linear velocity tracking & $\exp(-5\|\mathbf{v}_{xy}^{\text{cmd}} - \mathbf{v}_{xy}\|^2)$ \\
$r_{\text{ang}}$ & Angular velocity tracking & $\exp(-5(\omega_z^{\text{cmd}} - \omega_z)^2)$ \\
\midrule
\multicolumn{3}{l}{\textit{Negative Rewards (to be minimized)}} \\
$r_{\text{slip}}$ & Foot slip & $\sum_i \mathbbm{1}_{c_i>0} \cdot \|\mathbf{v}_{\text{foot},i}^{xy}\|^2$ \\
$r_{\text{smooth-1}}$ & Action smoothness (1st-order) & $\|p_t - p_{t-1}\|^2$ \\
$r_{\text{smooth-2}}$ & Action smoothness (2nd-order) & $\|p_t - 2p_{t-1} + p_{t-2}\|^2$ \\
$r_{\text{ori}}$ & Orientation deviation & $(\cos^{-1}(R_{zz}))^2$ \\
$r_{\text{pos}}$ & Joint position regularization & $\|W_q (q - q_0)\|^2$ \\
$r_{\text{vel}}$ & Joint velocity regularization & $\|W_{\dot{q}} \dot{q}\|^2$ \\
$r_{\text{acc}}$ & Joint acceleration regularization & $\|W_{\ddot{q}} (\dot{q}_t - \dot{q}_{t-1})\|^2$ \\
$r_{\tau}$ & Torque regularization (actuated joints) & $\|W_{\tau} \cdot \tau\|^2$ \\
$r_{\text{base}}$ & Base motion regulation & $0.4 v_z^2 + 0.2|\omega_x| + 0.2|\omega_y|$ \\
$r_{\text{contact}}$ & Body contact penalty & $c_{\text{body}}$ \\
$r_{\text{com}}$ & COM XY offset (standing mode only) & $\|\mathbf{x}_{\text{com}}^{xy}\|^2 \cdot \mathbbm{1}_{\text{stand}}$ \\
\bottomrule
\end{tabular}
\vspace{1ex}
\begin{minipage}{\linewidth}
\scriptsize
\textit{Notation:} 
$\mathbf{v}_{xy}$ and $\omega_z$ are base linear and yaw angular velocities; $(\cdot)^{\text{cmd}}$ denotes commanded values.  
$q$, $\dot{q}$, $\tau$ are joint position, velocity, and torque; $q_0$ is the nominal joint position.  
$W_q$, $W_{\dot{q}}$, $W_{\tau}$ are diagonal weight matrices.  
$p_t$ is the target joint position at time $t$.  
$R_{zz}$ is the $(3,3)$ entry of the base orientation matrix.  
$c_i$ is the contact state of the $i$-th foot; $c_{\text{body}}$ counts body-ground contact points.
\end{minipage}

\vspace{1ex}
\begin{minipage}{\linewidth}
\scriptsize
\textit{Reward Computation:} The standard reward is computed as:
\begin{equation*}
r_{\text{standard}} = \exp(0.2 \cdot r_{\text{neg}}) \cdot r_{\text{pos}},
\end{equation*}
where
\begin{equation*}
r_{\text{pos}} = \sum_i \alpha_i r_i^{\text{pos}}, \quad 
r_{\text{neg}} = \sum_j \beta_j r_j^{\text{neg}},
\end{equation*}
with $\alpha_i > 0$ and $\beta_j < 0$ representing positive and negative reward weights, respectively.
\end{minipage}
\end{threeparttable}
\end{table}

\begin{figure*}[!t]
  \centering
  \includegraphics[width=2\columnwidth]{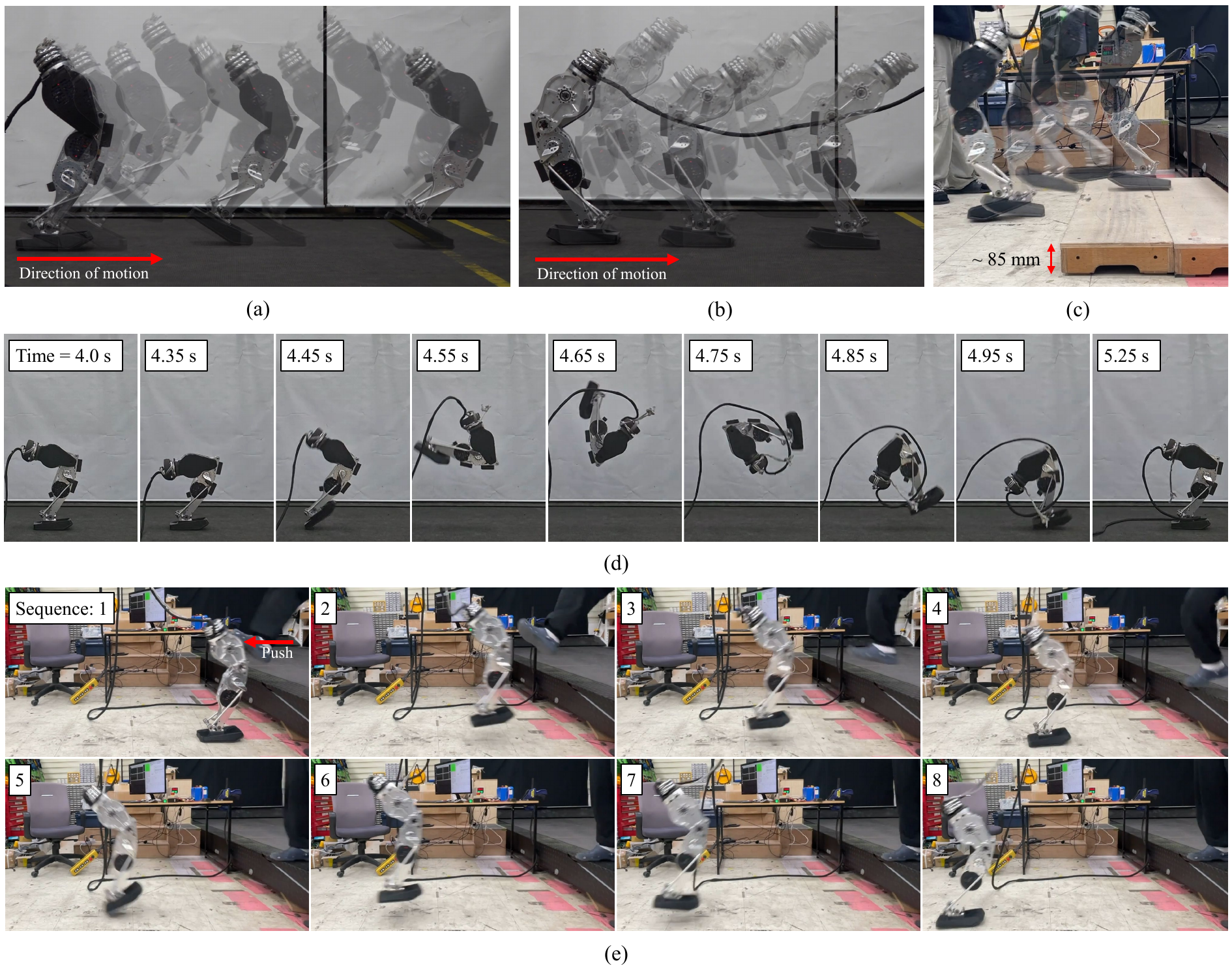}
  \caption{Snapshots from the experiments performed on the proposed 3-DOF hopping robot: (a) repetitive forward hopping, (b) repetitive backward hopping, (c) step-up, (d) front flipping, and (e) push recovery.}
  \label{fig:experiments_snapshot}
\end{figure*}

\section{Reinforcement Learning-based Control} \label{Sec:Control}

The locomotion controller is trained via reinforcement learning to track commanded linear and yaw rotational velocities using only proprioceptive observations.
We extend the previous reinforcement learning-based controller from~\cite{kim2024learning}, which combines a barrier-based reward for motion guidance and a standard regularization reward, to accommodate the proposed one-leg hopping robot.
The desired hopping frequency and foot clearance are predefined and enforced through the barrier-based reward to induce periodic motion.
To prioritize stable hopping, velocity tracking constraints are relaxed by widening the constraint boundaries.
Due to the flat-foot design, the robot can remain stationary even in a one-legged configuration, activating standing mode based on the command velocity~\cite{kim2024learning,lee2020learning}. 
However, because the support polygon is smaller in standing mode than in humanoid or quadruped robots, lateral deviations of the center of mass from the foot center are penalized. 
The barrier reward and standard reward are defined in Tables~\ref{tab:reward_barrier} and~\ref{tab:reward_standard}, respectively, and are separately processed by multiple critics~\cite{kim2024learning,kim2024not,zargarbashi2024robotkeyframing} and optimized using Proximal Policy Optimization (PPO)~\cite{schulman2017proximal}.

The ankle forms a closed-loop structure, with two D110A actuators mechanically coupled to generate pitch and roll motion, which must be considered in control.
We simulate this structure directly using RaiSim’s support for kinematic chains~\cite{hwangbo2018per}, where closed-loop systems are modeled by adding a pin constraint to a kinematic tree to enforce positional consistency between two points on separate links. 
Since each episode starts from randomly initialized generalized coordinates, the closed-loop may initially be inconsistent. 
To resolve this, we introduce a sub-step phase in which PD control is applied to the ankle’s pitch and roll joints to preserve their initial values, while the pin constraint gradually aligns the remaining links.
Once the configuration becomes sufficiently consistent, the episode begins. 


We add the joint position and velocity of the ankle to the privileged state, as they typically require forward kinematics and Jacobians to compute at deployment. These quantities are estimated from observations via a separate estimator and used by the actor. 
We adopt a concurrent learning setup~\cite{ji2022concurrent} with an asymmetric actor-critic architecture~\cite{nahrendra2023dreamwaq,pinto2017asymmetric}: the estimator predicts the privileged state, and the critic accesses it during training. 
The privileged state also includes body linear velocity, foot contact states, and local terrain height around the feet. 
The observation includes proprioceptive information with a short history, a cyclic phase function, and a standing-mode indicator.
The action defines target actuator positions for a PD controller.
All networks for the actor, critic, and estimator use Multi-Layer Perceptron models (MLPs).
Please refer~\cite{kim2024learning} for details on the observation.  

\section{Experiments} \label{Sec:Experiments}

To validate the dynamic capabilities of the proposed 3-DOF hopping robot, several actual hardware experiments were conducted.
All experiments were performed without any external support except for the umbilical cable.

\begin{figure}[!t]
  \centering
  \includegraphics[width=\columnwidth]{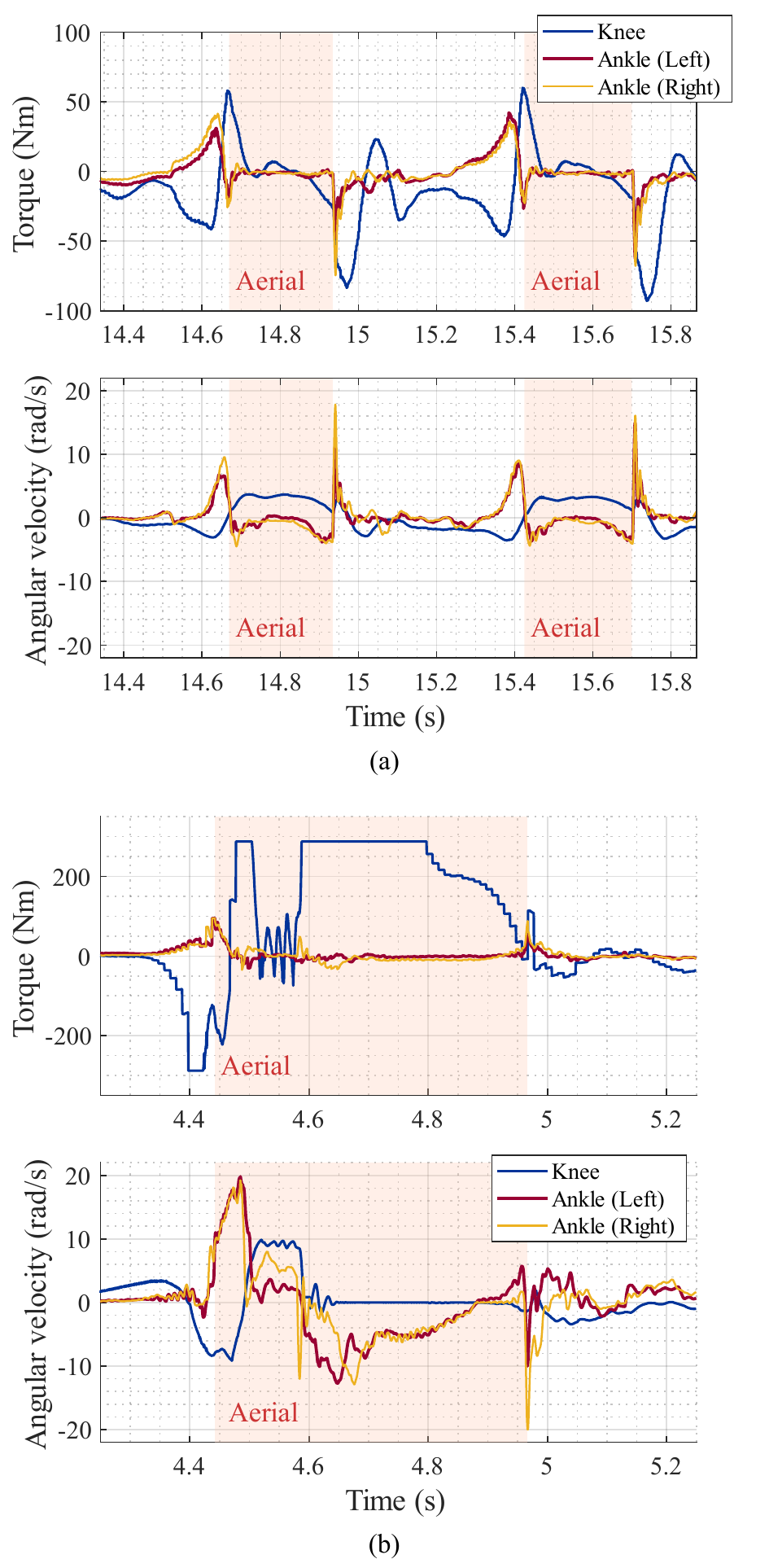}
  \caption{Measured joint torques and angular velocities during the experiments: (a) repetitive backward hopping motion, and (b) front flipping motion.
  The aerial phases are highlighted with orange-shaded regions.}
  \label{fig:experiments_data}
\end{figure}

As shown in Fig.~\ref{fig:experiments_snapshot}~(a) and (b), the robot successfully performed repetitive hopping motions in both forward and backward directions. 
Joystick commands were used to control the direction of motion, and the desired robot velocity was set to 0.8~\si{\meter/\second} in each direction.
The torque and angular velocity profiles of each joint actuator recorded during the backward hopping experiment are shown in Fig.~\ref{fig:experiments_data}~(a), where the aerial phases are highlighted with orange-shaded regions. 
The measured torque and angular velocity profiles show periodic patterns, confirming the robot’s ability to generate repetitive motions.

The policy for executing the front-flipping motion is designed based on the method proposed in~\cite{kang2025learning}.
The take-off was triggered by the user's joystick command, and the behavior is shown in Fig.~\ref{fig:experiments_snapshot}~(d). 
The robot successfully completed a full rotation and landed upright. 
The corresponding joint torque and angular velocity profiles, shown in Fig.~\ref{fig:experiments_data}~(b), indicate that the actuators generated high output to achieve the motion. 
In particular, the knee actuator reached both its torque and velocity limits, demonstrating that the actuation system was utilized to its maximum capability.

Two additional experiments were conducted to evaluate robustness against environmental changes and external disturbances.
As shown in Fig.~\ref{fig:experiments_snapshot}~(c), the robot successfully stepped onto an approximately 85~\si{\milli\meter}-high platform.
The same controller used for repetitive hopping was employed, and the motion was executed under user command. 

The result of the push recovery experiment is shown in Fig.~\ref{fig:experiments_snapshot}~(e). 
In this test, the robot was disturbed by an external force applied by a human foot, which triggered repetitive hopping motions. 
The robot successfully recovered its balance and continued operation without falling. 
This result shows that the robot possesses sufficient disturbance rejection capability to maintain stability under sudden perturbations, with its flat-foot design and reliable actuation hardware.

Experimental results show that the proposed 3-DOF hopping robot is capable of performing repetitive hopping motions as well as dynamic maneuvers such as front flipping.
It also showed robustness to external environmental changes, including step terrains and external disturbances. 
Additionally, these results verify that the proposed platform possesses the actuator capability to execute high-torque and high-velocity demanding dynamic maneuvers, along with sufficient structural strength to withstand impact forces during aggressive motions. 
In this regard, the proposed hopping robot can serve as an effective intermediate platform for the development of bipedal robots.
\section{Conclusion and Future Work} \label{Sec:Conclusion}

In this paper, a 3-DOF hopping robot was designed and fabricated as an intermediate platform toward bipedal robots capable of performing dynamic motions, and its operational performance was validated through actual hardware experiments.
From a hardware perspective, all major components—including the motor, gearbox, motor driver, and EtherCAT–CAN converter—were custom-designed to satisfy the system’s torque and angular velocity requirements as well as structural constraints.
In particular, to achieve a high reduction ratio while maintaining a hollow shaft structure within the 3K compound planetary gearbox, a MINLP-based gear teeth optimization method was proposed and applied to the gearbox design.
From a control perspective, a full-body simulation environment was constructed, fully incorporating the nonlinear and complex dynamics arising from the closed-loop link structure used for ankle power transmission.
Based on this simulation, a reinforcement learning-based controller was trained, which successfully enabled the robot to perform dynamic motions such as repetitive hopping and front flipping, while stably responding to environmental changes, including step terrains and external disturbances.

As future work, a bipedal robot is under development based on the electrical architecture and gearbox design methodology proposed in this paper. The robot is being designed with reference to the anthropometric data of an adult male (165~\si{\centi\meter} in height and 75~\si{\kilo\gram} in weight).
It is expected to achieve stable and dynamic walking, along with robust disturbance rejection in complex environments, by leveraging the high-torque performance of the custom-designed hardware, hollow-shaft-based wiring, and a reliable communication system.



\bibliographystyle{IEEEtran}
\bibliography{IEEEabrv,References}

\end{document}